\documentclass{article}
\usepackage[utf8]{inputenc}

\usepackage{amsmath}
\usepackage{amssymb}
\usepackage{graphicx}

\usepackage{booktabs}
\usepackage{multirow}
\usepackage{bbm}

\usepackage{amsmath}
\usepackage{amssymb}
\usepackage{mathtools}
\usepackage{amsthm}
\usepackage[capitalize,noabbrev]{cleveref}
\usepackage{bm}
\usepackage{multirow}
\usepackage{rotating}

\usepackage{geometry} 
\usepackage{longtable} 
\usepackage{cite}

\usepackage{multirow}
\usepackage{graphicx}
\usepackage{amssymb}
\usepackage{amsmath}
\usepackage{booktabs}
\usepackage{array}

\usepackage{amsfonts, amsmath, amsthm, amssymb}
\usepackage{graphicx}
\usepackage{booktabs}
\usepackage{wrapfig} 
\usepackage{cite}

\usepackage{multirow}
\usepackage{graphicx}
\usepackage{amssymb}
\usepackage{amsmath}
\usepackage{booktabs}
\usepackage{array}

\usepackage{graphicx} 
\usepackage{subcaption} 

\newcolumntype{L}[1]{>{\raggedright\let\newline\\\arraybackslash\hspace{0pt}}m{#1}}
\newcolumntype{C}[1]{>{\centering\let\newline\\\arraybackslash\hspace{0pt}}m{#1}}
\newcolumntype{R}[1]{>{\raggedleft\let\newline\\\arraybackslash\hspace{0pt}}m{#1}}

\usepackage{bbm}

\usepackage{lipsum}
\usepackage{changepage}
\usepackage{color,soul}
\usepackage{enumitem}   

\usepackage{pifont}

\usepackage{relsize}

\usepackage{tikz}

\makeatletter
\newcommand{\thickhline}{%
    \noalign {\ifnum 0=`}\fi \hrule height 1pt
    \futurelet \reserved@a \@xhline
}
\newcolumntype{"}{@{\hskip\tabcolsep\vrule width 1pt\hskip\tabcolsep}}
\makeatother

\usepackage{array}

\usepackage{framed,multirow}

\usepackage{latexsym}

\usepackage{url}
\usepackage{xcolor}

\usepackage{bm}

\usepackage{geometry}
\title{Streamline tractography of the fetal brain in utero with machine learning}

\author{Weide Liu$^1$, Camilo Calixto$^{1,2}$, Simon K. Warfield$^1$, and Davood Karimi$^1$ \\ $^1$Boston Children's Hospital and Harvard Medical School, Boston, MA \\ $^2$Elmhurst Hospital Center and Icahn School of Medicine at Mount Sinai, New York, NY}

\begin{document}

\maketitle

\begin{abstract}

Diffusion-weighted magnetic resonance imaging (dMRI) is the only non-invasive tool for studying white matter tracts and structural connectivity of the brain. These assessments rely heavily on tractography techniques, which reconstruct virtual streamlines representing white matter fibers. Much effort has been devoted to improving tractography methodology for adult brains, while tractography of the fetal brain has been largely neglected. Fetal tractography faces unique difficulties due to low dMRI signal quality, immature and rapidly developing brain structures, and paucity of reference data. To address these challenges, this work presents the first machine learning model, based on a deep neural network, for fetal tractography. The model input consists of five different sources of information: (1) Voxel-wise fiber orientation, inferred from a diffusion tensor fit to the dMRI signal; (2) Directions of recent propagation steps; (3) Global spatial information, encoded as normalized distances to keypoints in the brain cortex; (4) Tissue segmentation information; and (5) Prior information about the expected local fiber orientations supplied with an atlas. In order to mitigate the local tensor estimation error, a large spatial context around the current point in the diffusion tensor image is encoded using convolutional and attention neural network modules. Moreover, the diffusion tensor information at a hypothetical next point is included in the model input. Filtering rules based on anatomically constrained tractography are applied to prune implausible streamlines. We trained the model on manually-refined whole-brain fetal tractograms and validated the trained model on an independent set of 11 test scans with gestational ages between 23 and 36 weeks. Results show that our proposed method achieves superior performance across all evaluated tracts. The new method can significantly advance the capabilities of dMRI for studying normal and abnormal brain development in utero.

\end{abstract}

\section{Introduction}
\label{sec:introduction}

\subsection{Background and motivation}
\label{sec:background}

Medical imaging techniques have played an increasingly prominent role in understanding the development of human brain in utero \cite{malinger2004fetal, glenn2010mr}. Imaging modalities such as magnetic resonance imaging (MRI) are also becoming more popular in studying brain abnormalities prior to birth \cite{de2022adverse, hosny2010ultrafast}. As a consequence, fetal MRI has become an indispensable tool in medicine and neuroscience. Because the fetal period represents the most dynamic and most vulnerable stage in brain development, the potential medical and scientific impacts of fetal brain imaging are enormous.

Among various medical imaging modalities, diffusion MRI (dMRI) has emerged as a unique method for studying the fetal brain \cite{huppi2006diffusion, ouyang2019delineation}. It offers important insights into tissue microstructure and structural connectivity that no other imaging technique can provide. Research on adult brains has extensively documented that the microstructure of brain white matter tissue is a unique indicator of normal and abnormal brain development \cite{bodini2009diffusion, salat2014diffusion}. Moreover, structural connectivity of the brain can influence and be influenced by the progression of brain diseases \cite{collin2013ontogeny, fornito2015connectomics}. These findings suggest that dMRI can play a crucial role in studying the fetal brain in-utero, where it undergoes its most formative developments. With technical advancements in fetal imaging \cite{deprez2019higher, snoussi2024haitch}, an increasing number of studies have analyzed the normal and abnormal development of fetal brain via tract-specific assessment of white matter \cite{khan2018tract}. Moreover, there is growing interest in quantitative assessment of brain's structural connectome in the fetal period \cite{kasprian2013assessing, jakab2015disrupted}. Recent evidence suggests that the structural connectivity of the brain in utero can be altered by environmental factors and diseases. For example, prenatal exposure to maternal stress \cite{scheinost2017does}, congenital heart disease (CHD) \cite{schmithorst2018structural}, and brain malformations \cite{jakab2015disrupted} may disrupt the normal development of the structural connectome in utero.

Despite these unique potentials, dMRI-based assessment of the fetal brain has progressed very slowly. This is in large part because of a lack of reliable computational methods to automate the image data analysis. For analyzing adult brain dMRI data, there exists a repertoire of reliable computational tools \cite{theaud2020tractoflow, tournier2019mrtrix3, garyfallidis2014dipy}. For the fetal brain, similar tools are almost entirely nonexistent. This gap in technology has seriously limited our ability to tap the potential of dMRI to study brain's earliest and most critical developments. As an example, the only study to assess the impact of CHD on fetal brain white matter relied on manual tract delineation \cite{khan2018tract}. Because of the time and expertise requirements, that study was limited to one single tract and only a handful of subjects at one single gestational age. Automatic computational methods specifically tailored to fetal brain data can drastically enhance our ability to analyze larger fetal cohorts across the gestational age. Such tools can also reduce the cost and time requirements of these studies, increase the accuracy, and improve the reproducibility compared with manual annotations.

The goal of this work, therefore, was to develop and validate a new method for tractography of the fetal brain. Given the recent success of machine learning for this application \cite{poulin2019tractography, neher2017fiber}, our proposed method is based on a deep learning model that is trained on manually refined tractography data. If successful, such a method can dramatically improve the accuracy and reproducibility of quantitative fetal brain assessment with dMRI. It can enable precise automatic delineation of specific white matter tracts. Moreover, it can be used to reconstruct the structural connectome and to quantify the structural connectivity metrics in the fetal period.

\subsection{Related works}
\label{sec:related_works}

Bundle-specific and whole-brain tractography are widely used for delineating white matter tracts, for studying the white matter as a whole, and for reconstructing the structural connectome \cite{yeh2021mapping, zhang2022quantitative, lemkaddem2014global}. Early tractography algorithms relied on elementary models of the diffusion signal to compute the local fiber orientations and were primarily intended for qualitative or visual purposes \cite{basser2000vivo}. Over the past two decades, more advanced tractography algorithms have been developed \cite{smith2012anatomically, sepasian2012multivalued, li2014knowledge, yeh2019differential}. A representative example of these advancements is the class of global tractography algorithms \cite{mangin2013toward, christiaens2015global, lemkaddem2014global}, which treat tractography as a global inverse problem. Additional non-local information, such as anatomical context and history of streamline propagation, are often used by these advanced methods to improve the accuracy and reduce the false positive rates. There have been multiple efforts by the dMRI community to rigorously assess the capabilities and limitations of modern tractography algorithms \cite{maier2017challenge, maffei2022insights}. Some of the persistent challenges in tractography include fiber crossings and bottlenecks, inherent limitation of the dMRI data for inferring the local white matter fiber orientations, and paucity of ground truth data to develop and validate these methods \cite{zhang2022quantitative, rheault2020common}.

Machine learning may hold the key to addressing some of the perennial issues in tractography. A primary advantage of machine learning methods is their ability to integrate diverse sources of information, such as spatial information and anatomical priors. Classical non-machine learning tractography techniques can also incorporate such auxiliary information. However, it would be highly challenging to design those methods such that they optimally leverage different inputs. For example, the utility of different inputs likely depends on the streamline being traced and the position along the streamline. Conventional tractography methods do not have the capacity to learn such complex relations. Modern machine learning models such as deep neural networks, on the other hand, can learn these relations from training data. They can optimize their prediction objective with respect to different inputs much more effectively in a unified framework. Consequently, recent years have witnessed a growing interest in using machine learning for tractography. Reviews of these works can be found in \cite{poulin2019tractography, karimi2024diffusion}. 

Compared with adult brain studies, in utero fetal brain tractography has been much less utilized. This has been mainly because of the challenges of fetal dMRI acquisition and analysis. Compared with adult dMRI data, fetal brain dMRI measurements have overall inferior quality due to such factors as lower signal to noise ratio and motion effects. Moreover, different white matter tracts emerge and develop rapidly over a short span of time during the second and third trimesters. Due to these challenges, most prior works on fetal tractography have focused on one or a few selected tract bundles and, yet, have reported low reconstruction success rates and limited accuracy in reconstructing the tracts in their complete spatial extent \cite{kasprian2008utero, mitter2015vivo, zanin2011white, takahashi2014development, kolasinski2013radial}. Despite these limitations, prior works have also demonstrated that tractography has a great potential for studying normal and abnormal brain development in utero and for characterizing brain pathologies prior to birth \cite{jakab2015disrupted, huang2009anatomical, wilson2021development, kasprian2008utero, mitter2015vivo, zanin2011white, mitter2015validation, xu2014radial}. Therefore, accurate and reproducible fetal tractography will not only serve quantitative assessment of structural connectivity, but it can also advance the field of fetal brain imaging in other important directions.

\subsection{Summary of the contributions of this work}
\label{sec:contributions}

This work presents the development and validation of the first machine learning approach to streamline tractography of the fetal brain. Our proposed method follows design principles that make it well suited to the challenging nature of fetal dMRI data. 

We compute the local orientation of white matter fibers using a diffusion tensor fit to the dMRI signal. This enables the method to be used with typical dMRI scans that may not be suitable for determining complex fiber configurations. We use a neural network model, consisting of convolution and attention modules, to encode a large spatial context around the current streamline tracing point in the diffusion tensor image. In addition to the current streamline tractography point, we will also include the diffusion tensor information at a hypothetical next point. We also encode the information about the position of the current streamline point in the brain mask in terms of the distance to selected keypoints in the brain cortex. Moreover, similar to several existing methods, our model input includes the directions of recent streamline propagation steps. The tissue segmentation map is also encoded and fed to the model. Finally, we precisely align a spatio-temporal atlas of major fixels to the subject's brain and use that as additional input to the model. Our model synergistically combines these information to predict the next propagation direction.

We train the model on a set of whole-brain tractograms that have been manually edited and verified by human experts. Subsequently, we test our proposed method on a set of independent fetal brain scans. Our quantitative evaluations and visual assessments by an expert show that the proposed method achieves better results than a standard method and a machine learning technique. A comparison of our results with prior studies that have attempted tractography of the fetal brain shows that our results are vastly superior in terms of the range of white matter tracts that can be reconstructed and in terms of tract coverage. Therefore, our method represents a significant stride in improving the capabilities of dMRI for studying the brain in utero. The code will be available at \url{https://github.com/liuweide01/MLFT}.

\section{Methods}
\label{sec:methods}

\subsection{Image data acquisition and processing}

This study used retrospective MRI data collected from fetuses scanned at Boston Children’s Hospital. Structural and diffusion MRI scans were collected from each fetus. The dMRI data was acquired at a single \emph{b}-value of 500 and included between 24 and 96 measurements for each fetus. The raw data were processed with a well-tested fetal-specific preprocessing framework \cite{marami2017temporal} to correct for fetal/maternal motion and align all dMRI data into a standard space.

We fitted a diffusion tensor model to the data in each voxel. We then converted the diffusion tensor in each voxel to the diffusion orientation distribution representation \cite{descoteaux2008high} and expressed it in spherical harmonics of order 8. As a result, the fiber orientation information in each voxel was represented with a vector of length 45. The reason for conversion from the diffusion tensor to diffusion orientation distribution was to facilitate the interpolation during tractography and to avoid the interpolation errors that may arise when working with tensors \cite{zhang2006deformable}.

We used multi-atlas segmentation techniques to generate tissue segmentation and gray matter parcellation maps for each fetus. For tissue segmentation, we used a manually segmented fetal diffusion tensor atlas \cite{calixto2023detailed}. For each fetus, we registered the tensor atlas to the subject using a diffusion tensor-based registration method \cite{zhang2006deformable}. For gray matter parcellation, we used a T2-weighted atlas of the fetal brain \cite{gholipour2017normative}. We registered the atlas image to the mean diffusivity (MD) image of the subject. In both cases, we registered the three atlases that were closest in gestational age to the subject. Subsequently, we applied a probabilistic label fusion method \cite{akhondi2013simultaneous} to compute the final tissue segmentation and gray matter parcellation maps for the fetal subject. The tissue segmentation process involves classifying brain tissues into white matter, cortical gray matter, subcortical gray matter, and cerebrospinal fluid (CSF). These segmentations are essential for initiating and constraining the tractography process, especially for identifying and reconstructing complex white matter tracts.

Afterward, an expert visually inspected the diffusion maps (MD and FA), tensor glyphs (to ensure the accuracy of the principal eigenvector's direction), and tissue segmentation maps for each subject, removing any poor-quality data or incorrect segmentation maps. A total of 73 subjects were deemed to have acceptable data quality, with 62 subjects allocated to the training set and 11 subjects to the test set.

To generate reference streamlines for model training, we applied a method based on streamline propagation using the iFOD2 algorithm \cite{tournier2010improved}. Using the tissue segmentation maps, rules of anatomically constrained tractography \cite{smith2012anatomically} were applied to the iFOD2-generated results to remove implausible streamlines. The final results were reviewed by an expert to ensure they did not include gross errors. Finally, whole-brain tractograms were reviewed by an expert to ensure that all major tracts were present and that no gross errors were included.

\subsection{Method design}

\begin{figure}[!htb]
    \centering
    \includegraphics[width=1.0\linewidth]{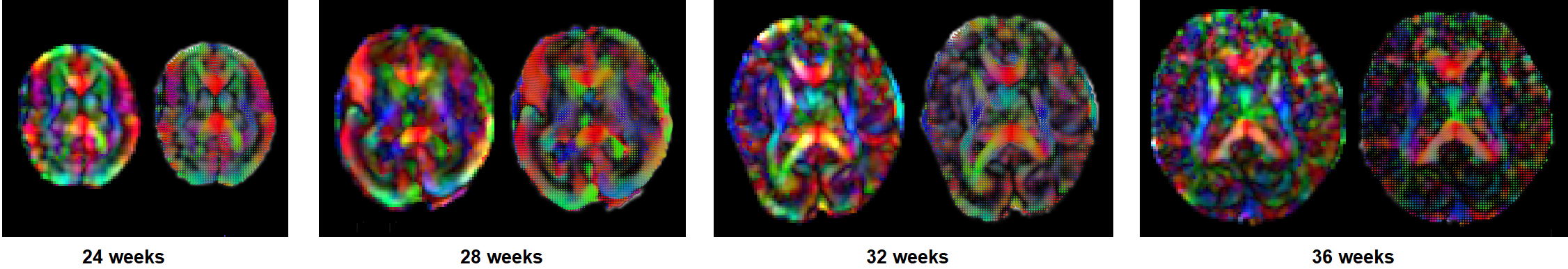}
    \caption{Example diffusion tensor and color fractional anisotropy images for fetal brains scanned at 24, 28, 32, and 36 gestational weeks.}
    \label{fig:example_tensors_and_cfas}
\end{figure}

Figure \ref{fig:example_tensors_and_cfas} shows maps of diffusion tensors and corresponding color fractional anisotropy images for four fetuses at 24, 28, 32, and 36 gestational weeks. These examples show that estimates of local fiber orientation can be very noisy and, on its own, not adequate for accurately guiding the streamline propagation.

To overcome this challenge, in this work we adopted several strategies. We briefly describe these strategies here and explain them in greater detail in the following subsections. First, we designed our method such that it leveraged the diffusion tensor information in a large spatial context around the current streamline propagation front. Second, we introduced additional local and non-local information to guide the streamline propagation. These information included recent propagation directions, local tissue segmentation map, and distance to standard keypoints in the brain cortex. Lastly, we used a spatio-temporal atlas of major fiber orientations, registered to the subject brain, to give the tractography algorithm the expected orientation of major fiber pathways. These strategies were meant to reduce the reliance on the estimate of local fiber orientation and enhance the robustness of the tractography results to the unavoidable errors in voxel-wise diffusion tensor fit.

\begin{figure*}[!htb]
\centering
\includegraphics[width=1.0\linewidth]{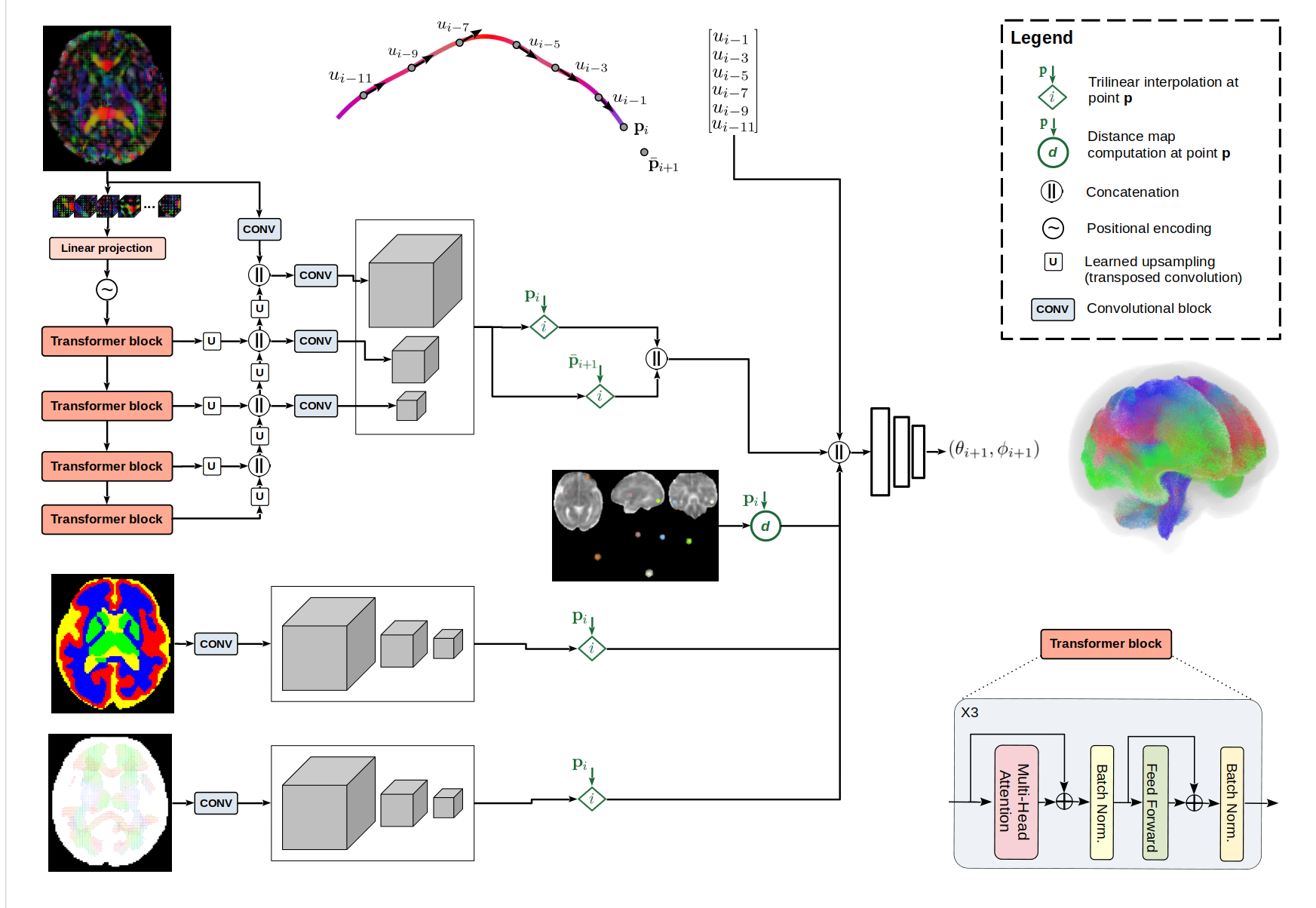}
\caption{The proposed fetal tractography method. The method encodes the information in the 3D volume of diffusion orientation distribution using transformer and convolutional blocks, generating feature maps at three different scales. The spatial location of the current streamline propagation point is used to interpolate these feature maps. A similar procedure is followed to encode the information in the tissue segmentation map and the fixel atlas image registered to the subject, although using a light-weight fully-convolutional network. These are combined with prior streamline propagation directions and with ``position vector'' features that represent the global location of the current point in the brain. A set of fully-connected layers fuse these features to predict the next streamline propagation direction.}
\label{fig:method_schematic}
\end{figure*}

Figure \ref{fig:method_schematic} shows the main components of the proposed method. Below, we first present the five sources of information that our method uses and explain how they are encoded. Subsequently, we describe how the method is trained and how it is applied to a test fetus.

\subsubsection{Fiber orientation information}

The largest computational unit in our method consists of an encoder-decoder deep neural network based on convolutions and transformer blocks that encode the fiber information from the input diffusion tensor image. The input to this unit is the entire $128^3$-voxel image of diffusion orientation distribution computed from the diffusion tensor in each voxel. The image is divided into a set of $8^3$ non-overlapping 3D patches. The patches are vectorized and projected into an embedding space. After positional encoding \cite{vaswani2017attention}, the sequence of embedded patches is processed by a set of 4 transformer blocks, each consisting of three transformer units with standard multi-head attention \cite{vaswani2017attention} and multilayer perceptrons. The design of the transformer blocks largely follows the vision transformer \cite{dosovitskiy2020image}. Skip connections involving transposed convolution operations provide shortcut paths between the encoder and decoder sections of the network. The output of the decoder includes feature maps at three different scales: \(F_1(H, W, D, C1)\), \(F_2(\frac{H}{2}, \frac{W}{2}, \frac{D}{2}, C2)\), and \(F_3(\frac{H}{4}, \frac{W}{4}, \frac{D}{4}, C3)\), where \(C1\), \(C2\), and \(C3\) denote the number of channel at each level. In this work, we set \(C1 = 64\), \(C2 = 128\), and \(C3 = 256\). The highest resolution feature maps, $F_1$, include features computed with a set of convolutional operations applied directly on the input diffusion orientation distribution image.

The feature maps thus computed provide a rich encoding of the diffusion orientation distribution image. Because of the deep and multi-scale design of the network, at each point in the image space the feature maps encode information about fiber orientation distribution from a large spatial extent around that point. For tractography, these feature maps need to be interpolated at the arbitrary location of the current streamline propagation point. We accomplish this via trilinear interpolation based on the voxels within a $3 \times 3 \times 3$ grid around the current point. Additionally, we use the values of the features at the centers of voxels within a $3 \times 3 \times 3$-grid around the current tractography point. To avoid a cluttered schematic, these last features are not shown in Figure \ref{fig:method_schematic}.

Additionally, we extracted features computed from the diffusion orientation distribution image at a hypothetical next point. Specifically, let us denote the current point with $\mathbf{p}_i$ and the direction of the last propagation with $\mathbf{u}_{i-1}$. We compute the hypothetical next point as $\bar{\mathbf{p}}_{i+1}= \mathbf{p}_i + \delta \mathbf{u}_{i-1}$, where $\delta$ is the look-ahead step size that we set to 0.5 voxels in all experiments. In essence, $\bar{\mathbf{p}}_{i+1}$ would be the next streamline point if the current propagation direction were to be the same as the last propagation direction, i.e., if we keep propagating in the same direction. Similar ideas have been advocated for in prior works. For example, Tournier et al. use a similar mechanism to obtain a look-ahead view of how the fiber orientation distribution looks like in front of the streamline being traced \cite{tournier2010improved}. We interpolated the feature maps extracted from the diffusion orientation distribution image at $\bar{\mathbf{p}}_{i+1}$ in a manner similar to that explained above for the current tractography point.

The features thus extracted from the diffusion orientation distribution image are concatenated and reshaped into a 1-D vector that is subsequently used, along with other inputs described below, by a multilayer perceptron (MLP) to compute the next propagation direction.

\subsubsection{Global location within the brain}

Another piece of information that we feed to the model consists of an encoding of the position in the brain mask. This information can be useful because similar patterns of local fiber orientation may exist in different parts of the brain and correspond to different white matter tracts. Encoding the global position of the current tractography point should be helpful in resolving such ambiguities and improving the reconstruction of different white matter tracts that may locally have similar shapes. Since the fetal brain grows dramatically, the challenge is how to achieve a consistent encoding of the position across the gestational ages. To address this challenge, we relied on brain parcellation labels. As mentioned above, we used the same parcellation scheme based on an existing spatio-temporal fetal brain atlas that covered all gestational ages \cite{gholipour2017normative}. We selected five specific parcellation regions from this atlas: (1) Left orbital part of the middle frontal gyrus, (2) Right insula, (3) Right inferior occipital gyrus, (4) Right middle temporal gyrus, and (5) Left inferior temporal gyrus.

\begin{figure}[!htb]
\centering
\includegraphics[width=1\linewidth]{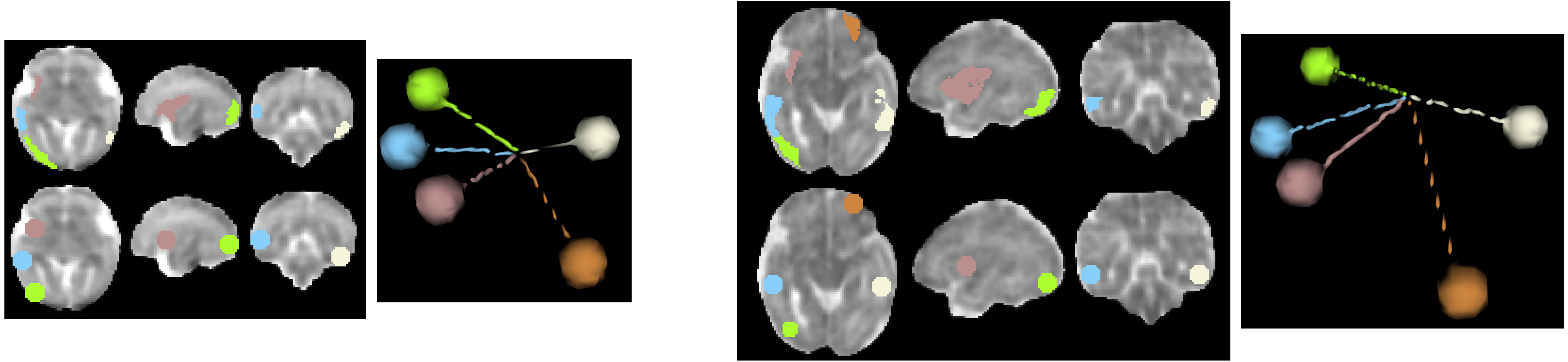}
\caption{Our proposed scheme for encoding the position of the current tractography propagation point in the brain. We encode this information as the normalized distance with respect to the centers of mass of five non-coplanar cortical parcellation regions. This figure shows the axial, sagittal, and coronal views depicting the cortical parcellation regions that are visible in the shown slices, spheres denoting the centers of mass of those regions, and 3D views that show the lines connecting these centers to an arbitrary point within the brain mask. The fetuses shown in this figure are 24 and 31 weeks of gestational age.}
\label{fig:distance_map}
\end{figure}

As shown in Figure \ref{fig:distance_map}, these regions were selected to be non-coplanar and far apart from each other in order to provide an unambiguous encoding of all positions within the brain mask. Let us denote the center of mass of each of these parcellation regions with $\{ \mathbf{c}_k \}_{k=1:5}$. We calculated the Euclidean distance from the current streamline position $\mathbf{p}_i$ to $\mathbf{c}_k$ as $r_k= \| \mathbf{p}_i- \mathbf{c}_k \|$ and form a normalized vector of these distances as the position encoding vector $\mathbf{r}_i$.

\begin{equation}
\mathbf{r}_i = \left[ \frac{r_0}{\sum_{k=0}^{5} r_k}, \frac{r_1}{\sum_{k=0}^{5} r_k}, \ldots, \frac{r_5}{\sum_{k=0}^{5} r_k} \right]
\end{equation}

This vector $\mathbf{r}_i$ is passed directly to the final MLP.

\subsubsection{Streamline propagation history}

An intuitively useful piece of information to guide the tractography is the streamline propagation history. This information has been used in some classical tractography algorithms as well. For example, Malcolm \textit{et al.} used an unscented Kalman filter to combine the streamline propagation history and the dMRI signal at the current streamline point to estimate the next propagation direction \cite{malcolm2010filtered}. Lazar \textit{et al.}, on the other hand, proposed a tensor deflection method that changed the propagation direction of an incoming streamline based on the shape of the diffusion tensor in the current voxel \cite{lazar2003white}. Leveraging the propagation history is also a common fixture of most machine learning-based tractography techniques. Poulin et al. \cite{poulin2017learn} and Jorgens et al. \cite{jorgens2018learning} directly incorporated the normalized vector of prior propagation directions as input to their model. Similarly, Neher et al. \cite{neher2017fiber} used the dot product of the prior direction as a weight for computing the next propagation direction in order to encourage alignment with the previous orientation. We have found that encoding the immediate prior propagation directions alone would make the model training difficult. Since most streamlines are locally very smooth, the immediate prior directions are often very close to the next target direction, making it difficult for the model to learn the small changes in direction between successive propagation steps. Hence, in our proposed method we skip some of the immediate prior steps and, instead, include earlier directions. Specifically, the propagation history that is used as input to our model includes $[u_{i-1}, u_{i-3}, u_{i-5}, u_{i-7}, u_{i-9}, u_{i-11}]$.

\subsubsection{Tissue segmentation map}

Streamlines are expected to start from brain gray matter, go through the white matter, and end in other gray matter regions. Therefore, segmentation of the brain tissue is highly informative for accurate tractography. Tissue segmentation maps are commonly used to reject anatomically invalid streamlines, for example, streamlines that prematurely stop inside the white matter or streamlines that cross voxels with cerebrospinal fluid (CSF). However, we think the tissue segmentation can be more effectively used to improve the accuracy of computing the propagation direction, rather than only for rejecting anatomically implausible streamlines. To this end, we use the tissue segmentation information in a small neighborhood around the current tractography point as input to our model.

We used a multi-atlas approach based on an existing diffusion tensor atlas of the fetal brain \cite{calixto2023detailed} to compute a tissue segmentation map for each fetal brain. We encode this information via a set of three learnable convolutional layers that compute three feature maps at different resolutions as shown in Figure \ref{fig:method_schematic}. We interpolate these feature maps at the position of the current streamline tractography point and pass the computed feature vectors to the final MLP.

\subsubsection{Spatio-temporal fixel atlas}

Due to the remarkable inter-subject similarity in brain structures, atlases are widely used to improve, constrain, and regularize the neuroimaging data analysis algorithms. They have also been used in a few prior studies on adult brain tractography. For example, Durantel et al. \cite{durantel2023riemannian} reconstructed a fixel atlas and used it as a prior to enhance the voxel-wise estimation of fiber orientation distribution (FOD). In each voxel, they computed an ``enhanced FOD" using a weighted combination of the FOD computed from the dMRI signal and the fixel prior, where voxel-wise metrics such as generalized fractional anisotropy were used to compute a heuristic weight. The enhanced FOD was then fed to a standard (non-machine learning) tractography method. A similar approach was followed by \cite{rheault2019bundle}, where the atlas (prior) and subject FODs were simply multiplied in each voxel. However, by using the atlas value in the current voxel only, these methods fail to fully exploit the information in the atlas.

\begin{figure}[t]
\centering
\includegraphics[width=0.6\linewidth]{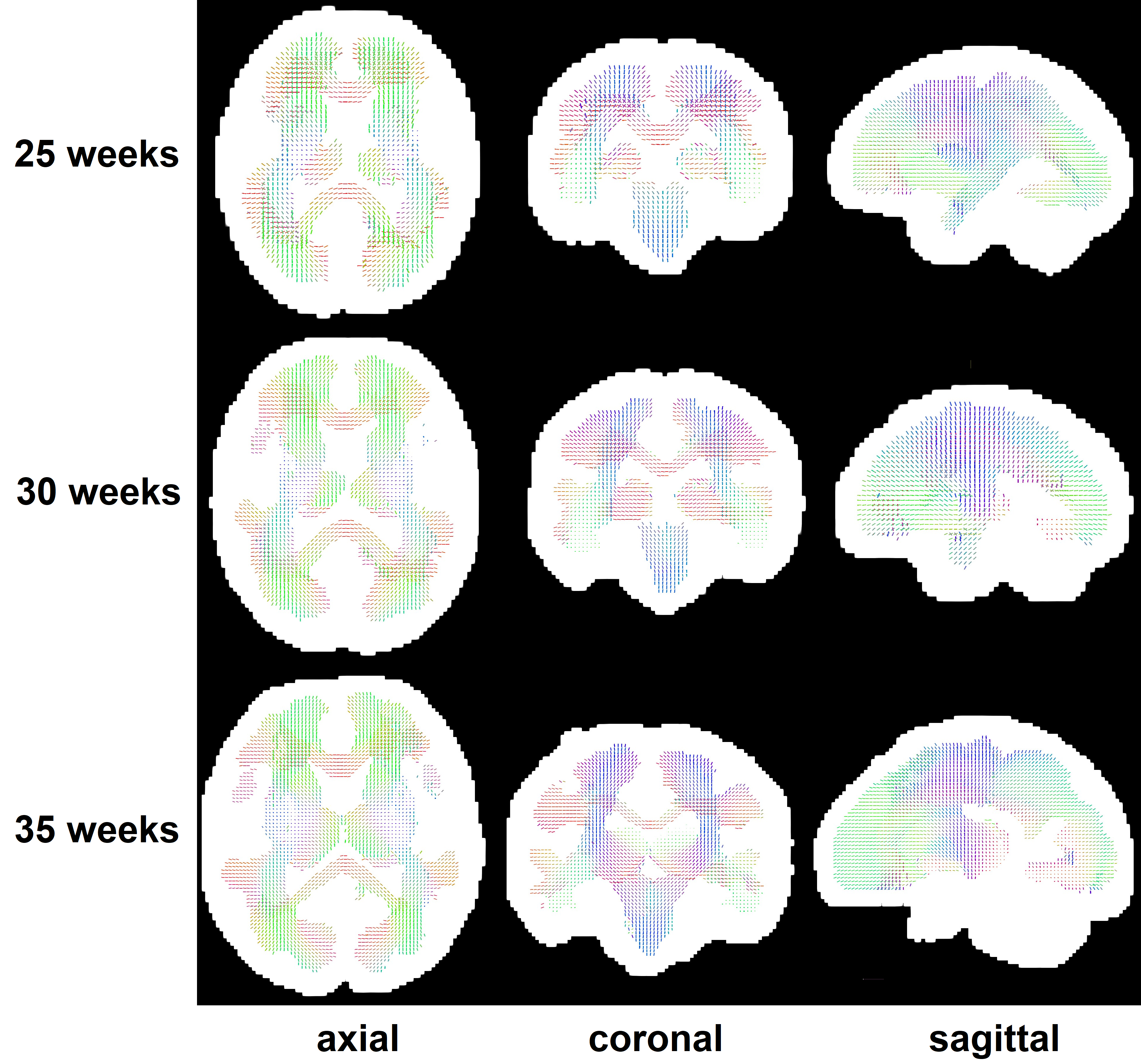}
\caption{Axial, coronal, and sagittal views of the spatio-temporal fixel atlas for 25, 30, and 35 gestational weeks. In order to better view the atlases for lower gestational ages, all atlases have been displayed to the same size.}
\label{fig:fixel_maps}
\end{figure}

In our method, we first reconstructed a spatio-temporal atlas of major white matter orientations in each voxel (i.e., fixels). The atlas was computed by applying these operations on the training data. (1) We computed whole-brain tractograms for each fetus. (2) Diffeomorphic diffusion tensor-based registration was used to align the tractograms for all fetuses of the same age into a common space. (3) Clustering algorithms were used to detect the directions of major fixels in each voxel based on the streamline directions in that voxel. Details of the method are described in \cite{calixto2024white}. Example fixel maps are shown in Figure \ref{fig:fixel_maps}. We found that only 2\% of the voxels contained more than two fixels. Therefore, in this work we used the directions of the first two fixels as additional input to our machine learning-based tractography model.

For each fetus, we register the age-matched fixel atlas using deformable diffusion tensor-based registration. As shown in Figure \ref{fig:method_schematic}, unlike prior works that have used a simple voxel-wise weighting of fiber orientations based on the atlas, we include the atlas information along with other information in our machine learning method. In other words, our proposed method computes the next streamline propagation direction by including the fixel atlas as one among many sources of information. We encode the fixel atlas information in a manner quite similar to the encoding of the tissue segmentation map. Specifically, we compute three sets of convolutional features at three different scales. We interpolate these feature maps at the location of the current tractography point and pass the computed feature maps directly to the final MLP for predicting the next propagation direction.

\subsection{Training procedures}

Our training dataset included a total of approximately 9 million streamlines from manually refined tractograms of 62 subjects. Although this is a notably large number, our preliminary experiments showed that a straightforward training on this dataset produced overly smooth streamlines at test time and did not allow the model to reconstruct the tracts with more varying curvatures. A simple data augmentation approach to increase the diversity of the target propagation direction resolved this issue. To apply this augmentation, we assumed the streamline direction $u$ followed a von Mises-Fisher (vMF) distribution $u \sim p(u; \mu, \kappa)= C(\kappa) \exp(\kappa \boldsymbol{\mu}^\top \mathbf{u}$). Here, $\mu$ and $\kappa$, respectively, denote the mean direction and concentration parameter of the distribution and $C$ is the normalizing coefficient. This is a standard model for representing distributions on the sphere and has been used in prior tractography works as well \cite{wegmayr2021entrack}. Minimizing the negative log likelihood of the observed direction gives rise to a simple loss function of the form $-\kappa \cdot \langle u, \hat{u} \rangle - \log C$, where $\langle ., . \rangle$ denotes the inner product. This formulation encourages the closeness of the predicted orientation with the target orientation while allowing for learned weighting of data with high uncertainty via modulating the loss term with $\kappa$. In other words, for data with higher uncertainty (larger $\kappa$) the model is allowed to make larger prediction errors by paying a higher ``uncertainty penalty''. In the second loss term, $C$ is a function of $\kappa$ and penalizes high uncertainties. This prevents degenerate training where the model computes high uncertainty for all data samples to reduce the first loss term.

While $\kappa$ may also be predicted by the model via including it in the loss function as explained above, in this work we use a heuristic approach to setting its value. Specifically, we set $\kappa$ to be proportional to the square of fractional anisotropy: $\kappa = \alpha \text{FA}^2$. We empirically set $\alpha$ to 1600. Most white matter voxels in our data had an FA value in the range [0.05, 0.25], resulting in $\kappa \in [4,100]$. For voxels with FA$>$0.20, which mostly represent voxels that are on major white matter tracts with little partial volume effect, $\kappa>64$. With the concentration factor $\kappa=64$, the 90\% confidence interval around the mean direction in the vMF distribution is a cone of approximately $12^\circ$. For a voxel with FA=0.10, which corresponds to smaller or less developed tracts or borders of major tracts with substantial partial volume effect, $\kappa= 0.16$ and the cone of 90\% confidence interval around the mean direction has an angle of approximately $24^\circ$. Regions with higher FA generally represent the location of the major tracts with a single dominant fiber orientation, where the streamline propagation direction is less ambiguous. Regions with lower FA, on the other hand, generally represent the voxels with low dMRI signal anisotropy due to significant partial volume effects or crossing fibers. Streamline propagation direction in these areas is less certain, which justifies increasing the value of $\kappa$. Having fixed $\kappa$ based on FA, we use the cosine similarity between the predicted and target propagation directions as the loss function.

\subsection{Implementation details}

The proposed method was implemented using PyTorch and trained on an NVIDIA A6000 GPU. We used a large batch size of 16,000 to improve the training stability. We used a Stochastic Gradient Descent optimizer with an initial learning rate of $1e-3$. Training was stopped when the training loss plateaued or began to fluctuate. In our experiments in this work, this usually happened after approximately 100 training epochs.

\subsection{Application of the trained model on a test scan}

\begin{figure}[t]
\centering
\includegraphics[width=0.6\linewidth]{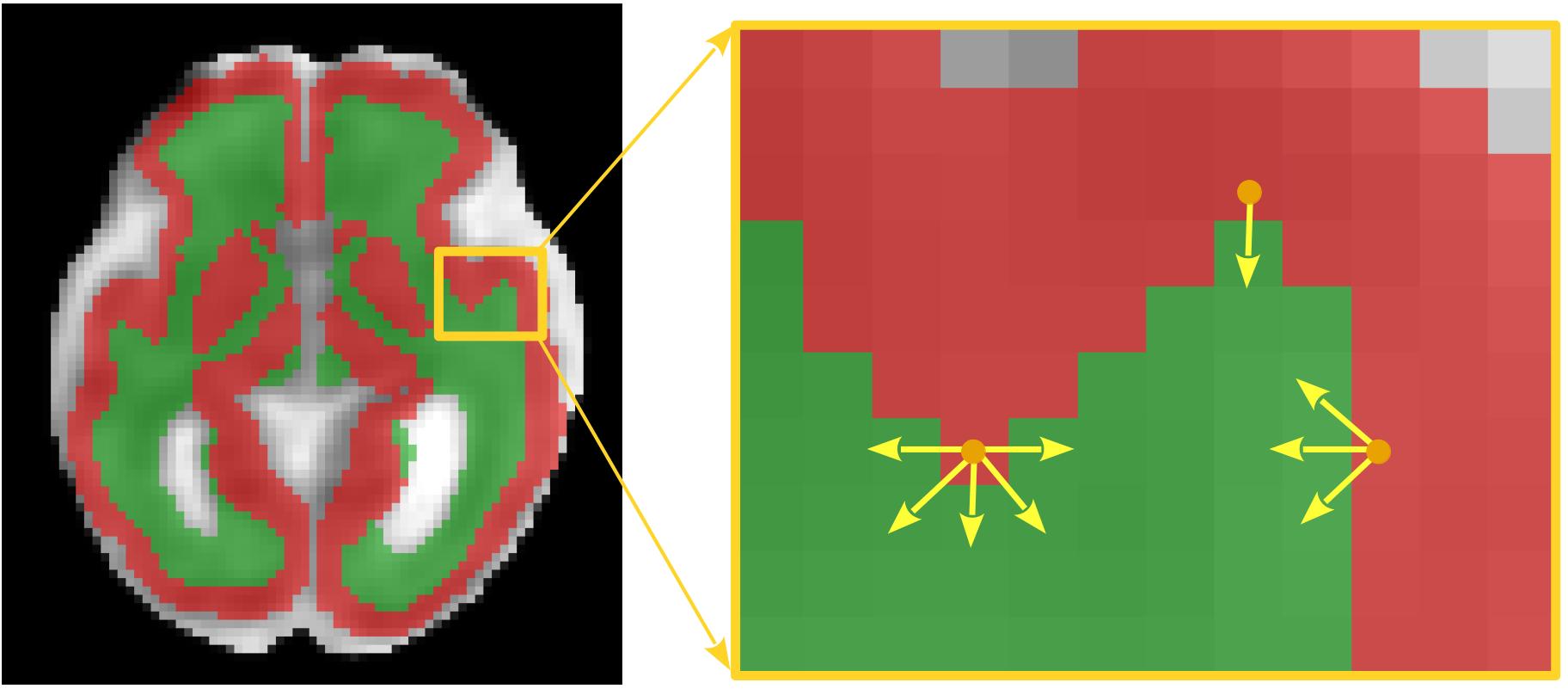}
\caption{We launch a streamline from the center of each gray matter voxel that has at least one white matter voxel neighbor. The direction of the first step is selected to be the direction of the line connecting the center of the gray matter voxel to the center of the neighboring white matter voxel. In this figure, the red voxels are gray matter and the green voxels are white matter. We have selected three arbitrary gray matter voxels and have shown the direction of the first step for the streamlines launched from those seed points with yellow arrows. To enhance the tractogram diversity, random jittering is applied to the location of the seed point and the direction of the first step as explained in the text. These augmentations are not portrayed in this figure.}
\label{fig:streamline_launch}
\end{figure}

Once the model training is complete, it can be applied to compute the whole-brain tractogram on a test subject. We use the tissue segmentation map for both streamline seeding and for rejecting implausible streamlines. Valid streamlines should begin and end at the boundary between gray matter and white matter. As shown in Figure \ref{fig:streamline_launch}, given the tissue segmentation map, we consider every voxel in the cortical gray matter that has at least one neighboring white matter voxel. We consider the vector between the centers of the gray matter voxel and that of the neighboring white matter voxel as the initial streamline propagation direction and launch a streamline in that direction. In order to increase the number and diversity of streamlines, we augment the position and direction of the streamline launch. We jitter the position of the seed point in the x, y, and z directions, uniformly, by [-0.6, 0.6]mm. We consider the initial direction in the spherical coordinate system with polar angle $\theta$ and azimuthal angle $\phi$, and add random values uniformly sampled from $[-30, 30]$ degrees independently to $\theta$ and $\phi$.

In order to enhance reconstruction of different tracts with varying shapes, we adopted a probabilistic tractography approach. Similar to the model training stage, we used a vMF distribution with the concentration factor modulated based on the local FA ($\kappa= \alpha \cdot \text{FA}^2$). Given the model's prediction of the next propagation direction $\mu$, we sample a random direction $\mu^{\prime} \sim \text{vMF} (\mu, \kappa)$ and propagate the streamline along $\mu^{\prime}$ by one step size. We use a step size of 0.6 mm, which is equal to half the voxel spacing in our data, in all experiments. We found that no single value of $\alpha$ resulted in optimal reconstruction results for all tracts. While the setting $\alpha= 1600$ was quite adequate for reconstructing tracts with large maximum local curvature such as some sections of the corpus callosum, it was not the best setting for some other tracts with lower curvature such as the corticospinal tract. Therefore, on a test scan, we apply the method with three different values of $\alpha \in [1600, 3200, 6400]$, to improve the reconstruction of various tracts. The final whole-brain tractogram for a test fetus is formed by simply merging the streamlines obtained with these three settings.

After propagating the streamline along $\mu^{\prime}$ by one step size, the model is applied again to compute the next propagation direction. This process is continued as long as the next streamline point is inside the white matter segmentation mask. The propagation is terminated once the predicted next point is not inside the white matter. If the next point is on a cortical or sub-cortical gray matter voxel, the computed streamline is considered correct and it is added to the tractogram. If the next point is on a CSF voxel or outside the intracranial mask, the streamline is considered as anatomically implausible and rejected. Additionally, streamlines that exceed a length of 130 mm are discarded, as they exceed the expected maximum streamline length for the fetal brain.

\subsection{Experiments and evaluation criteria}

Assessment and comparison of tractography algorithms is notoriously difficult. In this work, we used both quantitative metrics and qualitative assessment by a human expert to evaluate the proposed method and compare it with existing techniques. In order to reduce the subjectivity of the assessments, we used an automatic tool \cite{wassermann2016white} to extract a set of anatomically meaningful white matter tracts from the computed whole-brain tractograms. This method uses standard definitions of the tracts and a parcellation map of the gray matter to automatically extract specific tracts. In this study, we extracted nine tracts: Anterior Thalamic Radiation (ATR), Rostrum, Genu, and Splenium of the Corpus Callosum ($\text{CC}_1$, $\text{CC}_2$ and $\text{CC}_7$, respectively), Cortico-Spinal Tract (CST), Inferior Occipito-Frontal Fasciculus (IFO), Inferior Longitudinal Fasciculus (ILF), Optic Radiations (OR), and Uncinate Fasciculus (UF). Descriptions of these tracts can be found in \cite{calixto2024detailed, wasserthal2018tractseg}.

We compared our proposed method with a conventional non-machine learning method, Fiber Assignment by Continuous Tracking (FACT) \cite{mori1999three}, and a recent machine learning method based on recurrent neural networks (RNNs) \cite{cai2023implementation}. This RNN method was originally proposed for adult brain tractography and used a small spatial context of $3\times3\times3$ voxels for the model input. Because our preliminary experiments had shown that a larger spatial context could be useful in fetal tractography, we also applied a variant of the RNN method where we increased the input patch size to $7\times7\times7$ voxels. We refer to this method as ``modified RNN''.

Our quantitative assessments were in terms of the Dice Similarity Coefficient, precision, and recall of the reconstructed tracts compared with the ground truth. In order to compute these metrics, we needed to convert the streamline bundle for an extracted tract to a binary mask. To this end, we first computed the streamline density map and then removed the voxels where the streamline density was below the 5\textsuperscript{th}-percentile of the density values, which corresponded to spurious streamlines. To generate the ground truth binary tract masks for these evaluations, we applied the iFOD2 algorithm to compute 5,000,000 streamline for each fetus and extracted individual tracts using \cite{wassermann2016white}. An expert visually inspected each tract for every test fetus and marked the ones that were correctly and fully reconstructed. Only the tracts that passed the quality assurance were used in assessing our method and the compared methods in this work. 

The qualitative evaluation by the human expert was carried out by scoring the tracts on a 1-5 scale using a bespoke scoring system. The expert visually inspected the tract in 3D and assigned a score from 1 (lowest score, indicating ``failed reconstruction'') to 5 (highest score, indicating ``excellent reconstruction''). The scoring system is described in Table \ref{Table:expert_rating_scale}. In these assessments, the tracts reconstructed by different methods were presented to the expert in a random order, and the expert was blind with respect to which method had reconstructed the tract being assessed.

\begin{table*}[ht]
\centering
\caption{The scoring system used for qualitative assessment of the reconstructed tracts by a human expert in this study.}
\label{tab:tract_quality}
\begin{tabular}{clp{95mm}}
\toprule
\textbf{Score} & \textbf{Reconstruction Quality} & \textbf{Description} \\
\midrule
1 & Failed reconstruction & The method has failed to reconstruct any streamlines that could be considered as part of the expected tract. \\
\hline
2 & Poor reconstruction & The method has reconstructed some streamlines that correspond to the tract of interest. However, the streamlines substantially deviate into incorrect paths and they may be very sparse. Overall, it is difficult to correctly identify the tract based on the streamlines. \\
\hline
3 & Fair reconstruction & The reconstructed streamlines mostly follow the expected tract pathway, but there are notable deviations. The streamline paths and density, although overall plausible, suffer from small errors that lower its anatomical accuracy. \\
\hline
4 & Good reconstruction & The computed streamlines closely match the tract's anticipated anatomical path with high accuracy and sufficient density. Despite minimal errors, the streamlines are adequate to correctly define the tract's outline.  \\
\hline
5 & Excellent reconstruction & The streamlines correctly and completely match the expected tract pathway. They precisely cover the entire spatial extent of the tract. \\
\bottomrule
\end{tabular}
\label{Table:expert_rating_scale}
\end{table*}

\section{Results and Discussion}

Depending on the brain volume, approximately between 60,000 and 220,000 gray matter voxels with at least one neighboring white matter voxel existed for each brain. We launched five streamlines from each seed point, resulting in approximately between 300,000 and 1,100,000 streamlines per brain. Approximately 40\% of the computed streamlines were anatomically plausible based on the criteria described above, while the rest violated those criteria and were thus rejected. The retention rate of 40\% for our machine learning method is vastly higher than the retention rate of approximately 1\% for standard methods such as FACT that we have explored in this work. This suggests that the proposed method learns to compute more anatomically plausible streamlines, which not only improves the tractography accuracy but also reduces the computational time, the size of the tractogram, and the burden of subsequent tractogram post-processing.

\subsection{Quantitative assessments}

\begin{figure}[!htb]
    \centering
    \begin{subfigure}{0.32\textwidth}
        \centering
        \includegraphics[width=\linewidth]{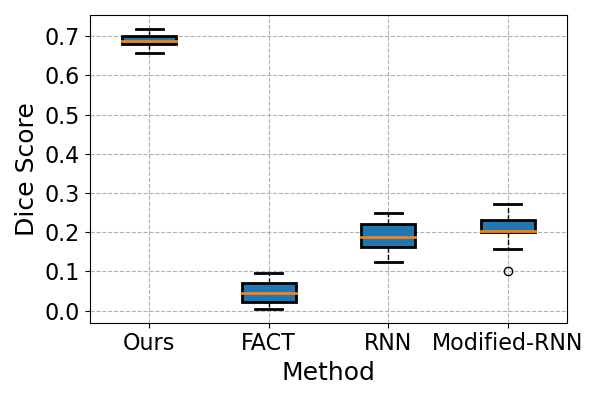}
    \end{subfigure}\hfill
    \begin{subfigure}{0.32\textwidth}
        \centering
        \includegraphics[width=\linewidth]{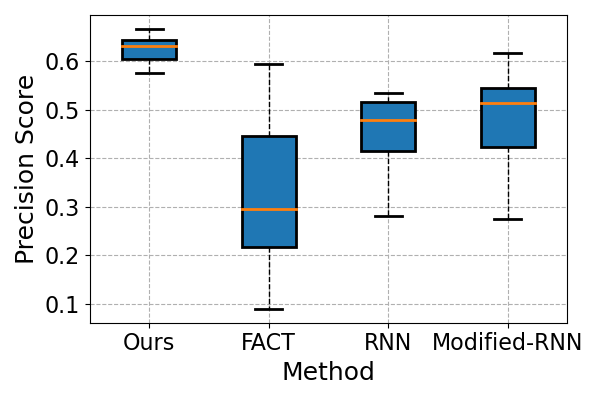}
    \end{subfigure}\hfill
    \begin{subfigure}{0.32\textwidth}
        \centering
        \includegraphics[width=\linewidth]{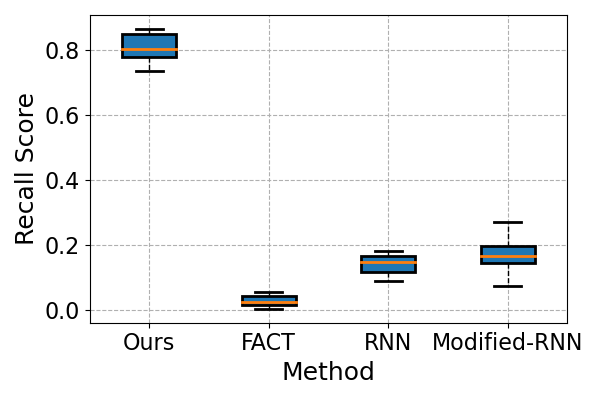}
    \end{subfigure}
    \caption{Summary of quantitative evaluation metrics for our proposed method and the compared techniques. The values shown in these plots have been pooled across 11 test subjects and 9 different white matter tracts.}
    \label{fig: dice_recall_precision}
\end{figure}

Figure~\ref{fig: dice_recall_precision} shows the results of quantitative evaluations in terms of DSC, precision, and recall. The plots in this figure show the summary of these metrics across all 11 subjects and all 9 white matter tracts. Our method achieved substantially higher DSC, precision, and recall than FACT and the two RNN baselines. While the mean and standard deviation of the DSC for our method was $0.711 \pm 0.027$, for FACT, RNN, and Modified-RNN, respectively, it was $0.048 \pm 0.028, 0.190 \pm 0.039,$ and $0.206 \pm 0.045$. Our method achieved a mean DSC of above 0.60 (across the 9 tracts) and a mean recall of above 0.80 for each of the 11 test fetuses that covered the gestational ages between 23 and 36 weeks.

\begin{figure}[t]
\centering
\includegraphics[width=1\linewidth]{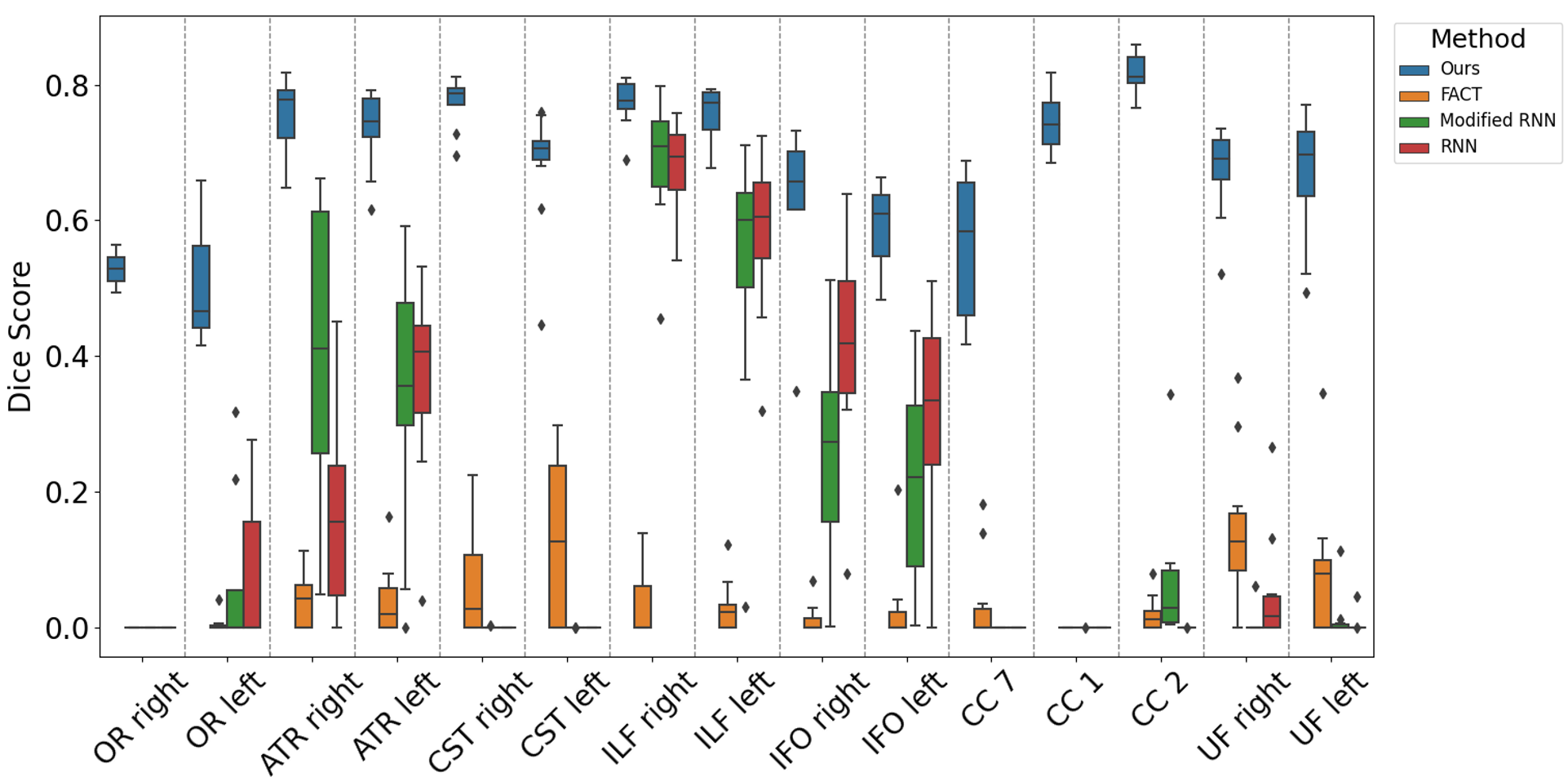}
\caption{Performance of our proposed method and compared techniques in terms of the Dice score for individual white matter tracts.}
\label{fig:results_tract_wise}
\end{figure}

Figure \ref{fig:results_tract_wise} shows the reconstruction accuracy for our proposed method and the compared techniques in terms of DSC, separately for each of the tracts. For the bilateral tracts, we have shown the results separately for the left and right tracts. This figure shows that the proposed method can achieve a mean DSC of close to 0.50 or higher for all tracts, which is highly satisfactory given the challenging nature of fetal tractography. Although a direct comparison of our results with experiments on adult brains is not possible, our results compare very favorably with recent studies on adult brain tractography \cite{poulin2022tractoinferno}. Nonetheless, the figure also shows significant variability in the reconstruction accuracy for different tracts. The proposed method has achieved mean DSC values of well above 0.70 for certain segments of the corpus callosum ($\text{CC}_1$ and $\text{CC}_2$), CST, and some of the association tracts such as ATR and ILF. For other tracts such as OR and the splenium of the corpus callosum ($\text{CC}_7$), on the other hand, the mean DSC is close to or below 0.60.

The variability in the reconstruction performance of the proposed method on different tracts may be attributed to at least three factors. First and foremost, being a machine learning technique, the performance of the proposed method is ultimately restricted by the accuracy of the training data. Our experience shows that the ability of the proposed method to reconstruct a tract depends directly on the presence and completeness of that tract in the whole-brain tractograms that were used in model training. As an example, compared with $\text{CC}_1$ and $\text{CC}_2$, the optic radiations (OR) was much less consistent in our training data. This has very likely contributed to the lower reconstruction accuracy for OR as shown in Figure \ref{fig:results_tract_wise}. As another example, the observed lateralization of tractography results in Figure \ref{fig:results_tract_wise} may be a manifestation of the impact of training data. We had noticed that the training tractograms used in this work were in general more accurate in the right hemisphere than in the left hemisphere. We speculated that this was almost certainly due to the lower accuracy in the left brain hemisphere of the tissue segmentation maps that were used in computing the whole-brain tractograms for the training data. Interestingly, as shown in Figure \ref{fig:results_tract_wise}, this has resulted in a consistently lower reconstruction accuracy for the left section of all the six bilateral tracts considered in this analysis. The impact of this factor may be ameliorated by improving the accuracy and reducing the bias of the training data.

The second factor that may have contributed to the differences in the reconstruction performance for different tracts may be the methodological choices and the algorithm settings. For example, as we have explained above, to generate the tractograms on test subjects we use a surface-based seeding approach, i.e. we launch streamlines on the boundary between the white matter and gray matter. Although this is in general a suitable seeding strategy \cite{st2018surface, zhang2022quantitative}, other seeding techniques may improve the reconstruction of certain tracts. Other settings in our method, such as the range of $\kappa$ values, are likely to benefit the reconstruction of some of the tracts more than others. A simple remedy to this problem may be to expand the range of parameter settings used to compute the tractograms, but this will also inevitably increase the computation time and may also give rise to higher false positive rates.

Lastly, the inherent ambiguities in streamline tractography may impact the reconstruction accuracy of different tracts to different degrees. Some of the main sources of these ambiguities include the difficulty of resolving complex fiber orientations based on the dMRI signal in each imaging voxel, fiber crossings, and bottleneck regions \cite{schilling2019limits, schilling2022prevalence}. These factors are known to impact tractography in general and have been extensively studied for adult brains, but have received little systematic treatment for fetal brains \cite{calixto2024white}. Overall, tracts such as $\text{CC}_1$ that go through fewer fiber crossing and bottleneck regions should be easier to trace compared with tracts such as OR that encounter such ambiguities more often through their course. Compared with the other two factors mentioned above, this third factor relates to the fundamental limitations of dMRI-based fiber tracing that may not be completely surmountable. Nonetheless, it has been suggested that machine learning methods such as the technique proposed in this work may represent one of the most promising approaches to addressing these limitations as well \cite{schilling2022prevalence}.

\subsection{Qualitative assessments}

\begin{figure*}[t]
    \centering
    \includegraphics[width=0.6\linewidth]{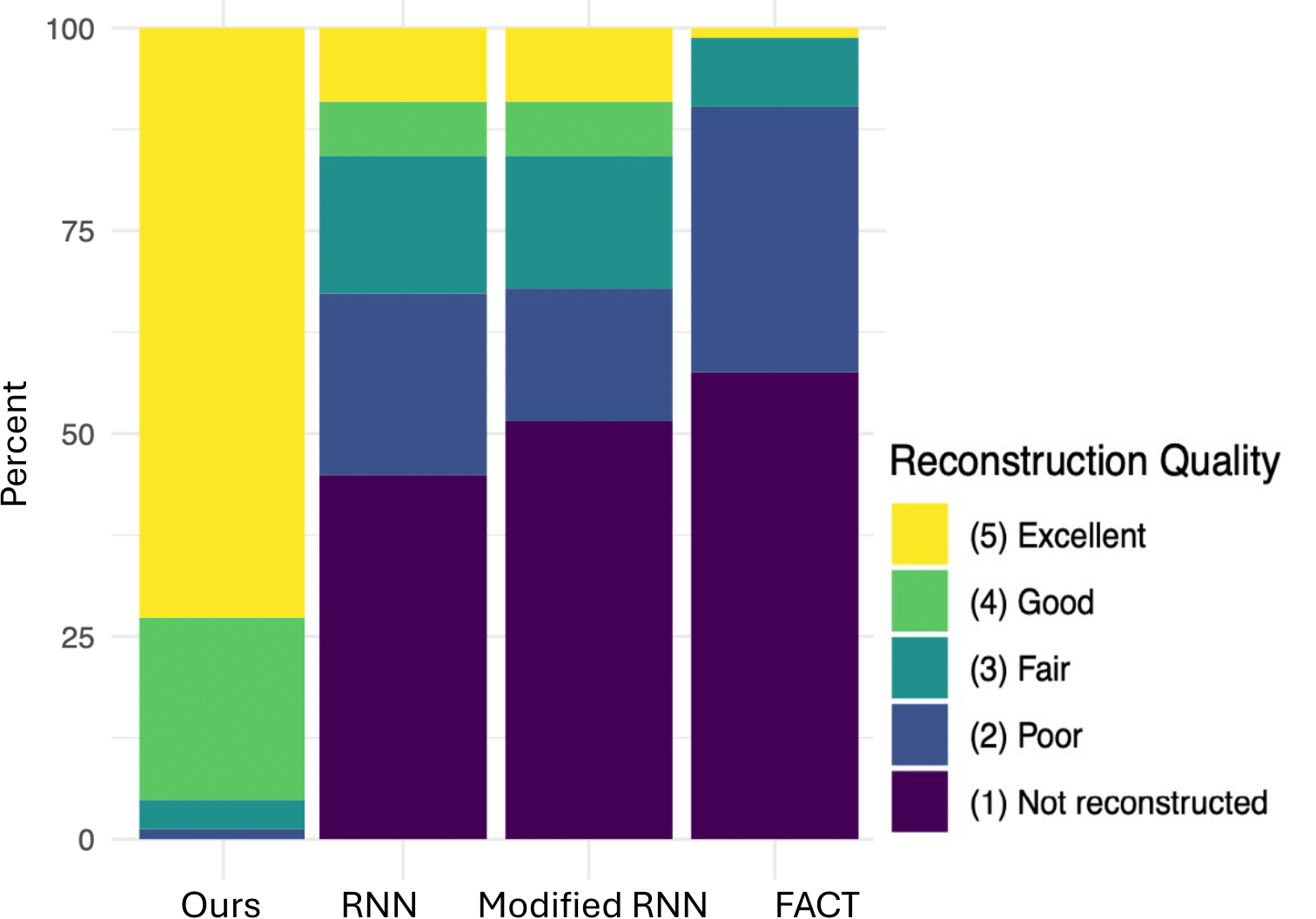}
    \caption{A summary of the quality scores assigned by an expert to the tracts reconstructed by the proposed method and the compared techniques. The results shown in this figure have been pooled across the 9 different tracts and the 11 test subjects. Descriptions of the scores are presented in Table~\ref{Table:expert_rating_scale}.}
    \label{fig:qua_score}
\end{figure*}

Figure \ref{fig:qua_score} shows the results of expert ratings based on the scoring system presented in Table~\ref{Table:expert_rating_scale}. This figure shows the summary of the scores across all 11 test subjects and all 9 white matter tracts considered in the assessments. On nearly 75\% of the cases, our method received the highest score (Excellent reconstruction). This means that in the great majority of the cases, the tract reconstructed by our method precisely matches the expected anatomy with optimal streamline density and almost no spurious streamlines. In most of the remaining cases, the tracts reconstructed by our method received the second highest score on the scale (Good reconstruction), suggesting that the reconstructed tracts had only minimal deviations from the expected anatomy and the streamline density was adequate to demarcate the tract's complete spatial extent. 

The scores assigned to the tracts reconstructed by the other methods showed almost an opposite pattern. The great majority of the tracts were either not reconstructed at all or received a ``Poor'' score. This means that, in an expert's opinion, the method either completely failed to reconstruct any valid streamlines or that the computed streamlines significantly deviated from the expected anatomy to the extent that it was not possible to identify the tract. This low performance was observed for the conventional method (FACT) as well as for the machine learning baseline (RNN). These observations show the inability of existing methods to adequately address tractography of the fetal brain and clearly point to the difficulty of this application.

\begin{figure*}[t]
    \centering
    \includegraphics[width=0.6\linewidth]{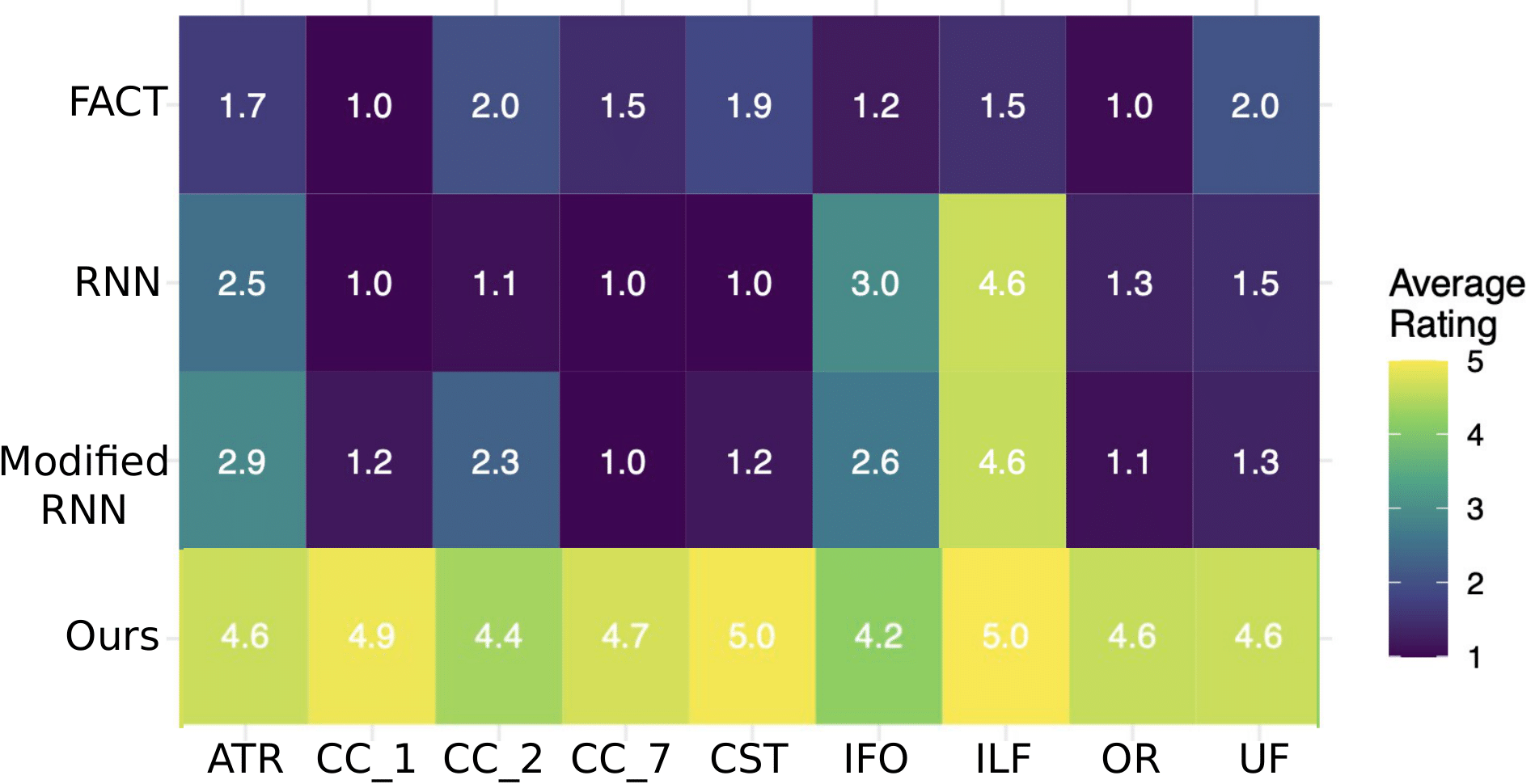}
    \caption{Detailed tract-wise reconstruction quality scores for the proposed method and the compared techniques. Descriptions of the scores are presented in Table~\ref{Table:expert_rating_scale}.}
    \label{fig:qua_detailed_score}
\end{figure*}

More detailed tract-specific scores are presented in Figure \ref{fig:qua_detailed_score}. They show that on all 9 tracts, the proposed method received an average score of above 4. This indicates that the method has accurately and faithfully reconstructed all tracts. RNN and Modified-RNN have accurately reconstructed the ILF and they have had moderate success in reconstructing ATR and IFO. However, they have performed very poorly in reconstructing the other six tracts. Similarly, FACT has failed to properly reconstruct any of the nine tracts. These results show that the proposed method can adequately reconstruct the full set of tracts considered in this study and that the existing techniques are clearly not well suited for this purpose.

A comparison of the results of this qualitative assessment with the quantitative assessment results presented in the above section can also be instructive. In particular, tracts such as OR and $\text{CC}_7$ that had received the lowest DSC scores in quantitative assessment have received high mean subjective scores of above 4.5 from the expert. This means that even though the overlap with the ground truth used for quantitative evaluation was not very high, in the expert's judgement the streamlines reconstructed by our method completely covered the expected anatomical extent of the tract. This may point to the unavoidable errors and inherent limitations of the ground truth used for quantitative assessments. More importantly, it suggests that, despite the relatively low quantitative evaluation metrics for some of the tracts, the streamlines computed by our method may be sufficient to serve some of the most important applications such as tract-specific analysis and structural connectivity.

\begin{figure}[t]
\centering
\includegraphics[width=0.6\linewidth]{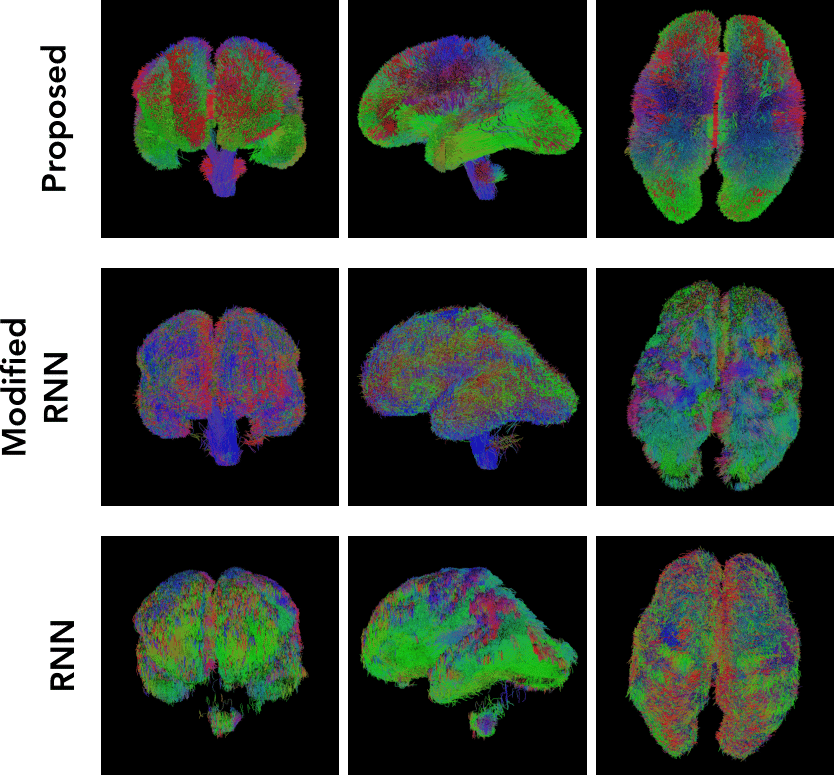}
\caption{Visual comparison of whole brain tractographies generated by our proposed method and other methods. Our method yielded the most favorable visual results, as can be seen in this depiction of a 26-week fetus.}
\label{fig:results_tractogram}
\end{figure}

\begin{figure*}[!htbp]
    \centering
    \includegraphics[width=0.99\linewidth]{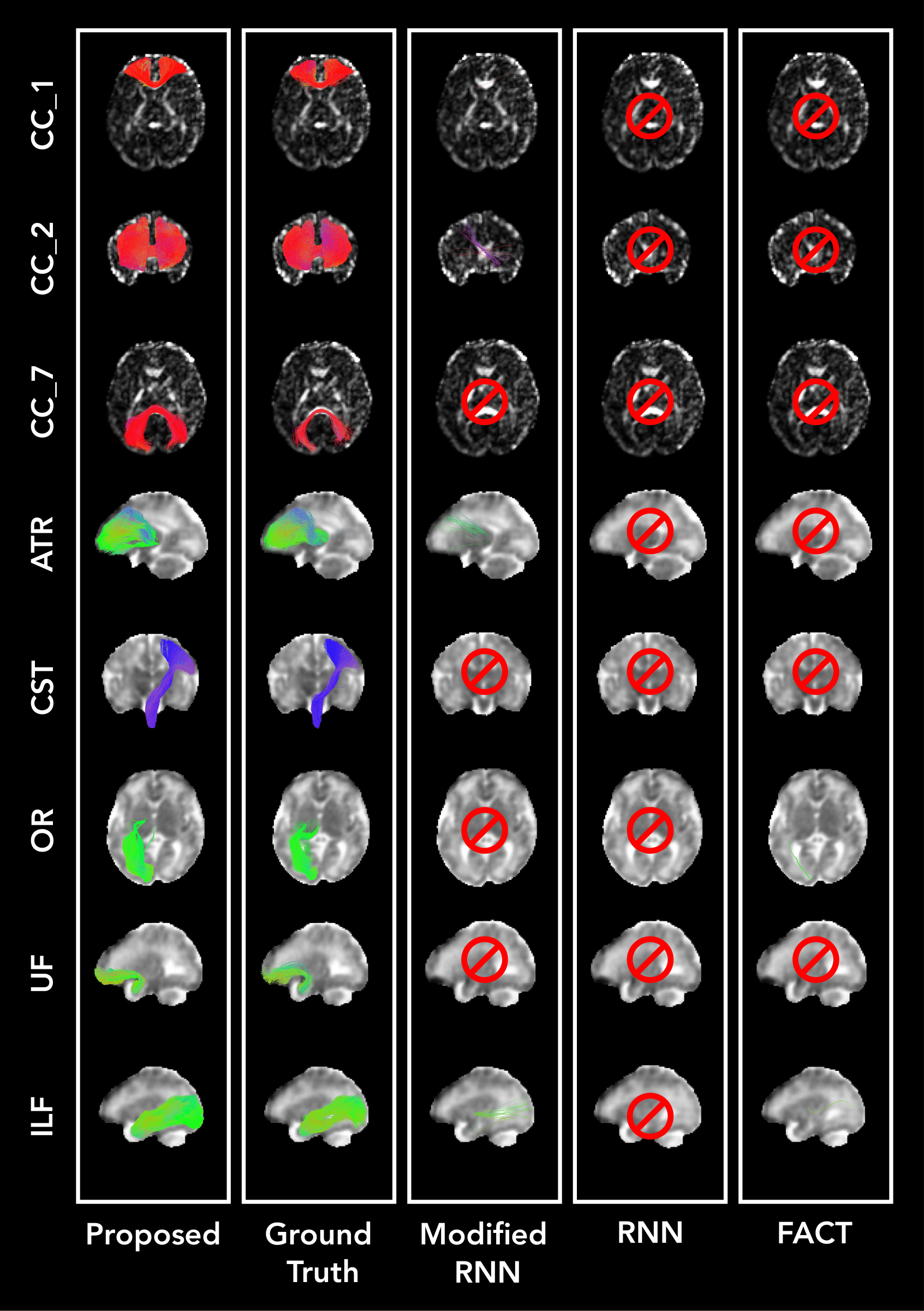}
    \caption{Comparison of tract reconstructions between our proposed technique and other methods on a 32-week fetus. In this example, our method reliably reconstructed all tracts, whereas other methods only managed to reconstruct a few tracts with low quality. Of note, for this specific test sample, RNN failed to reconstruct any of the tracts.}
    \label{fig:vis_results}
\end{figure*}

More visual comparisons of the results produced by our method and the compared techniques are presented in Figures \ref{fig:results_tractogram} and \ref{fig:vis_results}. Figure \ref{fig:results_tractogram} portrays the whole-brain tractograms for a test fetus at 26 gestational weeks, whereas Figure \ref{fig:vis_results} depicts the individual tracts. It shows that the proposed method can successfully reconstruct a range of association tracts (ATR, ILF, OR, IFO and UF), commissural tracts ($\text{CC}_1$, $\text{CC}_2$, and $\text{CC}_7$), and projection tracts (CST). The RNN and Modified-RNN methods have fair to moderate success in reconstructing some of the association tracts. However, they completely fail in reconstructing any valid streamlines for most tracts. FACT, on the other hand, has performed even more poorly than RNN and has failed to satisfactorily reconstruct any of the tracts.

\subsection{Comparison with prior works}

The significance of the advancements enabled by our method can be better appreciated by comparing our results with those of prior works. To this end, Table \ref{tab:tractography_summary} summarizes several prior studies that have attempted streamline tractography of the fetal brain. For each study listed in this table, we have briefly described the methodology followed to generate the tractography, the specific white matter tracts studied (if applicable), and a summary of the results. As this selection shows, past efforts have had limited success in reconstructing major white matter structures in the fetal brain. Interestingly, some of these studies have imaged post-mortem fetal brain specimens (ex-vivo) with specialized scanners. Many of them have also resorted to manual placement of ROIs to guide the tractography. Yet, most of these studies have achieved far inferior results compared with the results obtained by our method in this work. This comparison further points to the challenging nature of fetal tractography and to the significance of the results obtained with the new method.

\begin{table}[!ht]
\small
    \centering
    \begin{tabular}{|p{0.18\linewidth}|p{0.37\linewidth}|p{0.10\linewidth}|p{0.22\linewidth}|}
    \hline
    \textbf{Reference} & \textbf{Methods} & \textbf{Tracts} & \textbf{Results} \\
    \hline
Huang et al. \cite{huang2009anatomical} & This study scanned postmortem fetal brains between 13 and 22 gestational weeks using 4.7T and 11.7T scanners. Tractography was performed using FACT with user-specified ROIs. The authors followed protocols designed for adult brains to specify ROIs for different tracts. & CC, UF, ILF, FX & The method was able to reconstruct the specified tracts, although no success rate metrics were reported. Most tracts were only partially reconstructed. \\ \hline
Kasprian et al. \cite{kasprian2008utero} & The work computed tractography on fetal brains, scanned in utero, between 18 and 37 gestational weeks. Tractography was performed using FACT with manual specification of ROIs.  & CC, CST, TC & Authors reported an overall success rate of 40\%. Most reconstructed tracts showed only partial coverage of the anatomical extent of the tract. \\ \hline
Wilson et al.~\cite{wilson2021development} & Tracts were defined using multiple ROIs, including seed regions and inclusion/exclusion zones to confine tracts within one hemisphere. Tracts were refined by removing spurious streamlines via microstructure-informed filtering. & CST, OR, ILF, CC. & Tractography successfully generated most of the tracts, though OR was not identifiable at the earliest gestational stage. Often the tracts demonstrated only partial anatomical coverage. \\  \hline
Song et al.~\cite{song2015asymmetry} & This study was performed on a dataset of 23 postmortem fetal brains, imaged with 3T and 4.7T scanners. A two-ROI approach was employed to segment the streamlines. 
 & FX, ILF, IFO, AF & Most reconstructed tracts showed only partial coverage of the anatomical extent of the tract. \\ \hline
Jakab et al. \cite{jakab2015disrupted} & A total of 20 normal and 20 abnormal fetal brains were scanned in utero between 22 and 36 gestational weeks. The authors used an approximate white matter segmentation mask for tractography seeding. A fourth-order Runge-Kutta method was used for tractography. & NA & The study successfully used whole-brain tractograms to study structural connectivity. However, it did not assess the tractography results in terms of individual tracts.   \\ \hline
This study & A machine learning model based on typical \emph{in utero} fetal scans. The model is validated on an independent dataset of 11 fetal scans between 23 and 36 weeks of gestation. & ATR, $\text{CC}_1$, $\text{CC}_2$, $\text{CC}_7$, CST, IFO, ILF, OR, UF & All tracts have been successfully reconstructed across the gestational age. Expert assessments gave all tracts a mean score of between ``Good" and ``Excellent". \\
    \hline
    \end{tabular}
    \caption{A summary of the methods and results for selected prior works on fetal brain tractography. Tract name abbreviations are as follows. CC: corpus callosum; UF: uncinate fasciculus; ILF: inferior longitudinal fasciculus; FX: fornix; CST: corticospinal tract; TC: thalamocortical tracts.}
    \label{tab:tractography_summary}
\end{table}

The success of the new method developed in this work can be attributed to several factors. First, the proposed machine learning model can synergistically combine different local and non-local information that are useful for tractography in a unified framework in the form of a deep neural network that can be optimized end-to-end. Even for fiber orientation, which is inherently a local information, our design effectively exploits the information in a large neighborhood around the current streamline tracing point. This reduces the impact of local fiber orientation estimation errors that can be significantly higher in fetal brain studies than in adult studies. Our method also effectively uses other sources of information that are difficult to incorporate into conventional tractography methods. It utilizes accurate tissue segmentation maps to launch the streamlines, guide the streamline propagation, and reject anatomically implausible streamlines. Another novel aspect of the proposed method was the incorporation of a streamline orientation prior in the form of a spatio-temporal fixel atlas that was precisely aligned to the subject brain. Lastly, we fed a long history of propagation directions, from up to 11 steps preceding the current point, as well as the distance to keypoints in the brain cortex to inform the tractography method of the location within the brain. These information were easily passed to the method via late fusion with other information at the final MLP model. The machine learning model was trained on tractography results from 62 fetal brain scans across the gestational ages considered in this work. Therefore, unlike conventional methods that compute the next streamline propagation direction based purely on the local data from the given scan, our method computes the next step based on what it has learned after being trained on millions of streamlines from tens of subjects. Hence, our method represents an entirely different approach to streamline tractography, where local, non-local, and population-level information can be used jointly to address this challenging task.

\section{Conclusion}
\label{sec:conclusion}

This study presented the first machine learning approach to fetal brain tractography. Given the challenging nature of dMRI-based tractography in general, and the limited reliability of local fiber orientation computations in the fetal brain in particular, advanced machine learning methods seem to be the natural solution to this problem. These methods are able to synergistically combine multiple sources of information to overcome the insufficiency of the local fiber orientation information. The method presented in this paper exploits the diffusion tensor image, propagation history, global spatial information, tissue segmentation, and a fixel atlas prior. We validated our model on an independent dataset consisting of 11 fetal scans across gestational ages between 23 and 36 weeks, demonstrating that our method outperformed existing techniques in all evaluated tracts. A comparison with the prior art on fetal tractography shows that the results obtained with our method are substantially better than those reported in the literature. Improved tractography accuracy offered by our method can facilitate the tract-specific analysis of normal and abnormal fetal brain development with dMRI and the assessment of structural brain connectivity in utero.

\section*{Acknowledgements}

This research was supported in part by the National Institute of Neurological Disorders and Stroke under award number R01NS128281; the Eunice Kennedy Shriver National Institute of Child Health and Human Development under award number R01HD110772; National Institutes of Health grants R01LM013608, R01EB019483, and R01NS124212, and the Office of the Director of the NIH under award number S10 OD0250111.  The content of this publication is solely the responsibility of the authors and does not necessarily represent the official views of the NIH.

\bibliographystyle{unsrt}
\bibliography{davoodreferences}

\end{document}